\documentclass[10pt,twocolumn,letterpaper]{article}

\usepackage{cvpr}              
\usepackage{algorithm}
\usepackage{algpseudocode}  
\usepackage{tabularx}
\usepackage{amsmath}
\usepackage{amssymb}
\usepackage{amsfonts}
\usepackage{booktabs}
\usepackage{multirow}
\definecolor{cvprblue}{rgb}{0.21,0.49,0.74}
\usepackage[pagebackref,breaklinks,colorlinks,allcolors=cvprblue]{hyperref}

\def\paperID{44830} 
\def\confName{CVPR}
\def\confYear{2026}

\title{Beyond the Golden Data: Resolving the Motion-Vision Quality Dilemma via Timestep Selective Training}
\makeatletter
\newcommand{\printfnsymbol}[1]{%
  \textsuperscript{\@fnsymbol{#1}}%
}
\renewcommand*{\@fnsymbol}[1]{\ensuremath{\ifcase#1\or *\or \dagger\or \ddagger\or
   \mathsection\or \mathparagraph\or \|\or **\or \dagger\dagger
   \or \ddagger\ddagger \else\@ctrerr\fi}}

\author{
    Xiangyang Luo$^{1,2}$\printfnsymbol{1} \quad Qingyu Li$^{2}$\printfnsymbol{2} \quad Yuming Li$^{2}$ \quad Guanbo Huang$^{1}$ \quad Yongjie Zhu$^{2}$ \\ Wenyu Qin$^{2}$ \quad Meng Wang$^{2}$ \quad Pengfei Wan$^{2}$ \quad Shao-Lun Huang$^{1}$\printfnsymbol{2}   \\
    $^1$Tsinghua University ~~ $^2$Kling Team, Kuaishou Technology
}

\begin{document}
\maketitle
{\let\thefootnote\relax\footnotetext{{$^{*}$ This work was conducted during the author’s internship at Kling Team, Kuaishou Technology.  

~~ $^{\dagger}$ Corresponding authors. }}}
\begin{abstract}
Recent advances in video generation models have achieved impressive results. However, these models heavily rely on the use of high-quality data that combines both high visual quality and high motion quality. In this paper, we identify a key challenge in video data curation: the Motion-Vision Quality Dilemma. We discovered that visual quality and motion intensity inherently exhibit a negative correlation, making it hard to obtain golden data that excels in both aspects. To address this challenge, we first examine the hierarchical learning dynamics of video diffusion models and conduct gradient-based analysis on quality-degraded samples. We discover that quality-imbalanced data can produce gradients similar to golden data at appropriate timesteps. Based on this, we introduce the novel concept of Timestep selection in Training Process. We propose Timestep-aware Quality Decoupling (TQD), which modifies the data sampling distribution to better match the model's learning process. For certain types of data, the sampling distribution is skewed toward higher timesteps for motion-rich data, while high visual quality data is more likely to be sampled during lower timesteps. Through extensive experiments, we demonstrate that TQD enables training exclusively on separated imbalanced data to achieve performance surpassing conventional training with better data, challenging the necessity of perfect data in video generation. Moreover, our method also boosts model performance when trained on high-quality data, showcasing its effectiveness across different data scenarios. 

\end{abstract}    
\section{Introduction}
\label{sec:intro}
\begin{figure}[t]
    \centering
    \includegraphics[width=\linewidth]{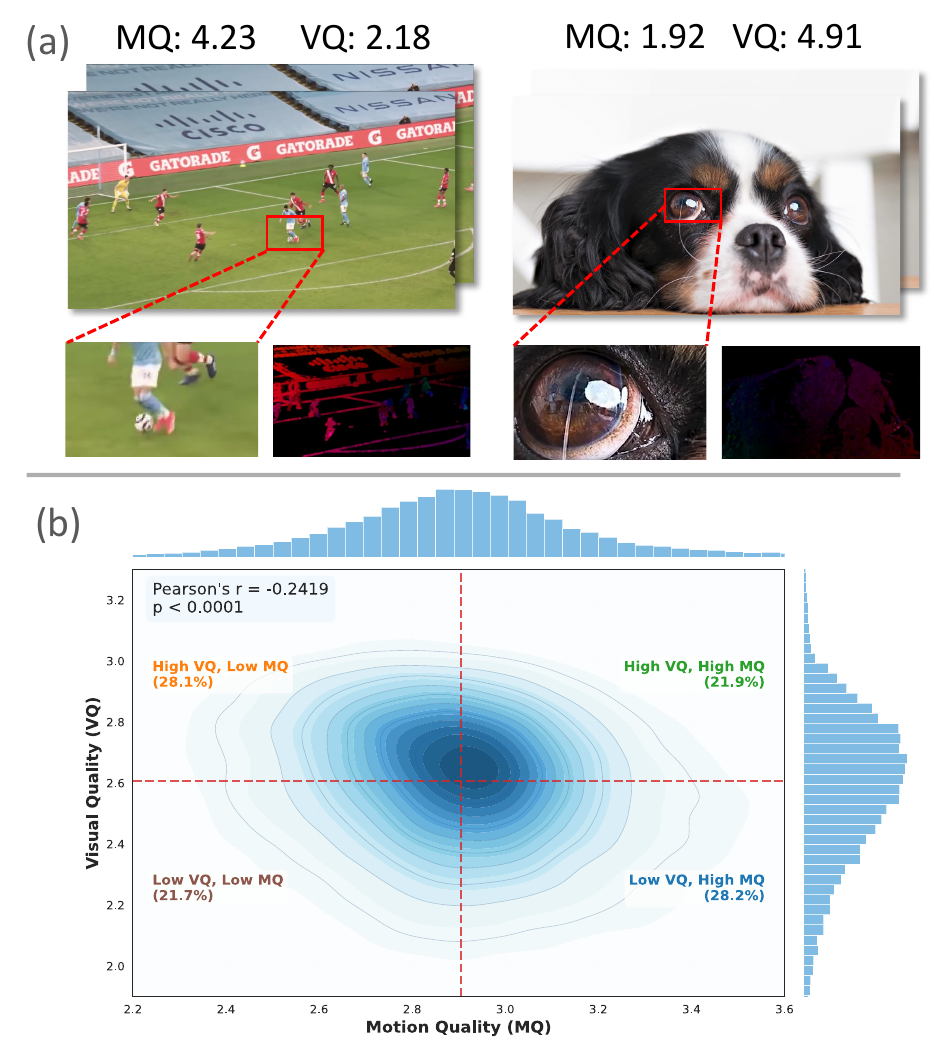}
    \caption{The motion-vision dilemma in video data. (a) Illustrates two cases: a high-MQ, low-VQ action scene (left) and a low-MQ, high-VQ static scene (right), with their corresponding optical flow maps below visualizing the difference in motion intensity. (b) The 2D density plot over Koala36M dataset confirms this trade-off. MQ and VQ exhibit a negative correlation, with the majority of samples ($56.3\%$ total) falling into the high-VQ/low-MQ or low-VQ/high-MQ quadrants, as defined by the median lines. Data with both high qualities is less common ($21.9\%$).}
    \label{fig:intro}
\end{figure}

Recent advancements in video generation models~\cite{animatediff,svd,ltxvideo,hunyuanvideo,wan} have made impressive strides in the field of Artificial Intelligence Generated Content. Video generation models are capable of producing high-quality, coherent videos, with applications spanning entertainment~\cite{movieagent, luo2025filmweaver, guo2026dreamid, luo2025canonswap, luo2024codeswap}, advertising~\cite{animegamer,human}, and education~\cite{education, isolate}. These models, however, heavily depend on high-quality training data—specifically, golden data that combines both high vision quality (VQ) and high motion quality (MQ). Unfortunately, such datasets are rare and expensive to obtain, which presents a major limitation in scaling these models.

Most existing works address this issue from the data selection perspective~\cite{huang2025d2c, select2, wan}. Large video datasets are first collected and then filtered using strict criteria to retain only samples that satisfy both high VQ and high MQ. For instance, videos with minor blur or irregular motion are typically filtered out, even though they may still contain valuable learning signals for either visual or motion representation. While this filtering improves training data quality, it also removes a substantial portion of potentially useful videos, leading to significant data waste and a dataset biased toward perfect samples.

To better understand the scarcity of such dual-quality data, we analyzed the Koala36M dataset~\cite{koala} and visualized the relationship between VQ and MQ, as shown in Figure~\ref{fig:intro}. Figure~\ref{fig:intro}(a) presents two representative imperfect cases: a high-MQ but low-VQ action scene (left) and a high-VQ but low-MQ static scene (right), with corresponding optical-flow visualizations highlighting their motion intensity differences. Figure~\ref{fig:intro}(b) further depicts the distribution of all samples, revealing a clear negative correlation between motion and visual quality ($r=-0.2419$). Only $21.9\%$ of videos fall into the quadrant with both high VQ and MQ (defined by median lines; practical filtering would use much higher thresholds), confirming that the golden data emphasized by prior data-selection pipelines is statistically rare. We refer to this inherent conflict between motion and visual fidelity as the Motion-Vision Quality Dilemma.

In this work, we take a different perspective—shifting the question from which videos should be kept to how can we use imperfect videos more effectively? We begin by analyzing the learning behavior of video diffusion models and make two key observations. First, the denoising process exhibits hierarchical learning: motion emerges at high-noise timesteps while details refine at low-noise timesteps.  Second, and more critically, through gradient-based analysis we discover that quality-imbalanced samples can match golden data's gradients at appropriate timesteps.
This reveals that imbalanced data need not be discarded but should be strategically allocated to matching denoising phases where they provide equivalent learning signals.
Building upon this observation, we adopt timestep selection in the training process and instantiate it through Timestep-aware Quality Decoupling. Instead of sampling timesteps equally across all data, we adjust the sampling distribution based on each video’s quality profile. High-MQ videos are biased toward higher-noise timesteps, whereas high-VQ videos are biased toward lower-noise timesteps. This design allows the model to leverage imperfect data effectively, aligning each sample with the phase of diffusion training where it provides the most benefit.

Extensive experiments demonstrate that TQD enables training exclusively on separated imbalanced data—containing only high-VQ/low-MQ or high-MQ/low-VQ samples—to achieve performance surpassing conventional timestep sampling trained on datasets with golden data, challenging the necessity of dual-quality samples in video generation. Moreover, TQD further boosts model performance when applied to high-quality data, showcasing its effectiveness and robustness across diverse data scenarios. These results open a more data-efficient and scalable path for video generation.
The key contributions of this paper are as follows:
\begin{itemize}
\item We identify and define the Motion-Vision Quality Dilemma in video generation.
\item We conduct gradient analysis revealing that quality-imbalanced data can produce learning signals equivalent to golden data at appropriate timesteps, providing theoretical foundation for timestep-selective training.
\item We introduce and implement Timestep-Selective Training and propose the Timestep-aware Quality Decoupling (TQD) framework for video diffusion models.
\item We empirically demonstrate that TQD allows training with imbalanced data to achieve performance surpassing conventional training method, validating the effectiveness of this new paradigm.
\end{itemize}

\section{Related Work}

\subsection{Video Diffusion Models}
Recent progress in video generation is largely driven by video diffusion models~\cite{wan,hunyuanvideo,yang2024cogvideox,ltxvideo,svd,grid,make,animatediff}, which synthesize videos by progressively denoising Gaussian noise into coherent temporal sequences. Early approaches~\cite{animatediff,svd} factorized spatial and temporal modeling through separated attention mechanisms—applying spatial self-attention within frames followed by temporal attention across frames. This decomposition enables efficient training by leveraging pretrained image models~\cite{pixart, stable, sd}, but the separated treatment of space and time can cause temporal artifacts and inconsistent motion. Recent architectures~\cite{openai2024sora,hunyuanvideo,yang2024cogvideox} have shifted toward unified 3D attention that jointly models spatiotemporal correlations, naturally capturing space-time dependencies and achieving better motion coherence, though at higher computational cost.

The training formulation has also evolved beyond the classical diffusion objective~\cite{ddpm}, which learns to predict noise $\epsilon_\theta(x_t, t)$ by minimizing $\mathcal{L}_{\text{diff}} = \mathbb{E}[\|\epsilon - \epsilon_\theta(x_t, t)\|^2]$, where $x_t = \sqrt{\bar{\alpha}_t}x_0 + \sqrt{1-\bar{\alpha}_t}\epsilon$. While this stochastic denoising excels at spatial fidelity, recent frameworks~\cite{hunyuanvideo,wan,opensora} adopt flow matching~\cite{flow}, which deterministically transports samples from noise to data by learning a velocity field: $\tfrac{dx_t}{dt} = v_\theta(x_t, t)$. This ODE-based formulation directly models probability flow, favoring temporally stable trajectories and smoother motion. Despite these advances, existing methods all assume high-quality data with both excellent visual and motion quality. Our work addresses the practical challenge of quality imbalance, demonstrating that principled timestep-aware training effectively leverages imperfect data and achieves consistent improvements across both diffusion and flow matching paradigms.

\subsection{Data Selection and Curation}

Data curation plays an important role in large-scale generative modeling. In vision-language modeling, both curated datasets~\citep{schuhmann2022laion5b} and recent model-driven selection approaches~\citep{feuer2024select,chen2024less,xie2024influence,li2024boosting} show that selecting informative examples under limited compute can significantly improve generalization.
In diffusion models, explicit data-centric work is more limited. D2C~\citep{huang2025d2c} condenses training sets via difficulty scoring; FiFA~\citep{yang2024fifa} filters preference data based on informativeness; and Alchemist~\citep{alchemist2024} upgrades noisy public text-to-image data through iterative filtering. Mild corruption applied to pre-training data has also been shown to improve robustness~\citep{chen2024slightcorruption}. Works like Ambient Diffusion~\citep{ambient} demonstrate that training on uncurated or imperfect data can yield robust models, highlighting the potential of data-centric approaches to handle noisy data.
For video generation, recent systems~\citep{wan,hunyuanvideo} adopt multi-stage filtering pipelines evaluating visual quality, motion consistency, and caption alignment. Yet these pipelines discard videos whenever either VQ or MQ is low, meaning that samples strong in only one dimension are never exploited. In contrast, our approach explicitly leverages such imbalanced data by routing high-MQ/low-VQ and high-VQ/low-MQ videos to different regions of the denoising process, reducing reliance on rare jointly high-quality video samples.

\subsection{Timestep and Loss Reweighting}

Diffusion models learn across a sequence of noise levels, and many works redesign timestep sampling or loss weighting to improve training. Foundational improvements such as cosine schedules~\citep{nichol2021improved} and SNR-based loss reweighting~\citep{hang2023minsnr} highlight that different timesteps contribute unequally to optimization. Perception Prioritized training~\citep{choi2022p2} further adjusts weights to favor perceptually important denoising stages, while curriculum-based strategies~\citep{curriculum} gradually expand the range of timesteps used during training to improve stability. Most of these methods apply a global strategy shared across samples, without explicitly adapting timestep allocation to the characteristics of individual training examples.

Recent work has also explored timestep-dependent data utilization beyond such global designs. In particular, Ambient Diffusion Omni~\citep{ambient} demonstrates that selectively utilizing imperfect or out-of-distribution data across diffusion timesteps can improve optimization, highlighting the importance of timestep selection during training.
Inspired by this line of research, we introduce timestep-dependent perspective into video training. Different from prior work that mainly studies general imperfect-data utilization, we identify the motion--visual quality dilemma in large-scale video data and propose Timestep-aware Quality Decoupling (TQD), which assigns sample-dependent timestep distributions according to motion and visual quality, enabling imbalanced video data to contribute more effectively during training.

\section{Method}

\begin{figure*}
    \centering
    \includegraphics[width=\linewidth]{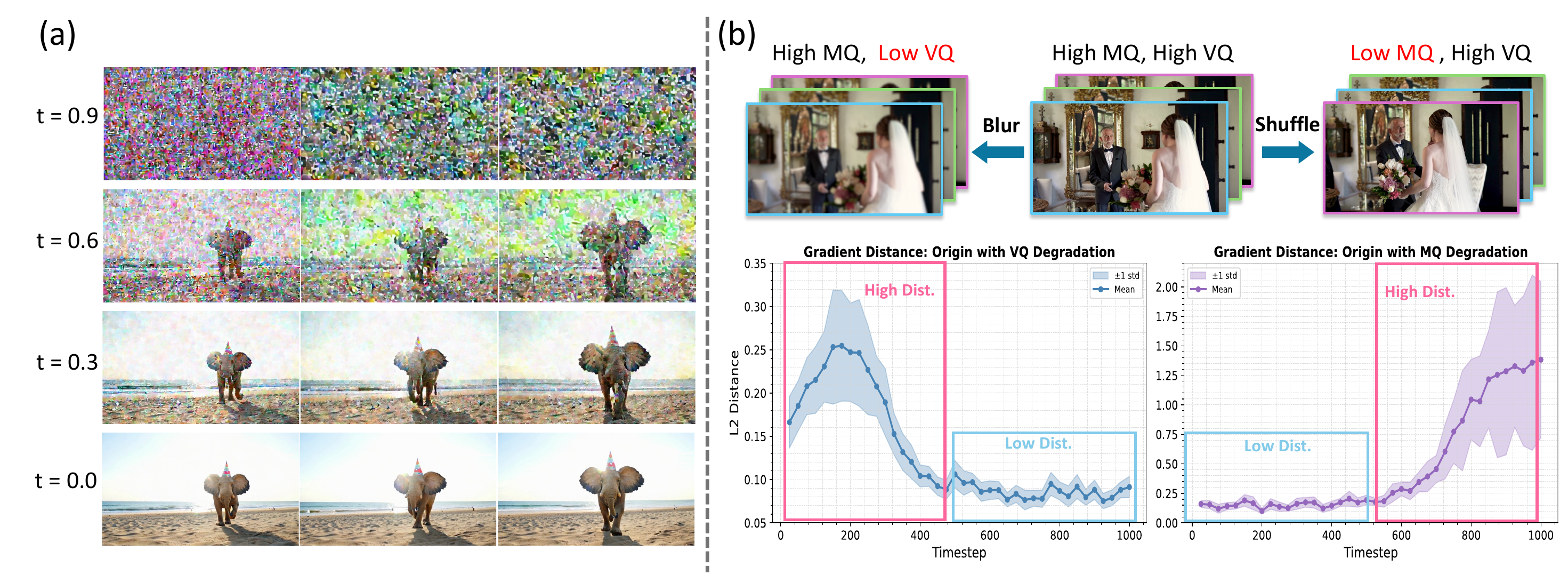}
    \caption{(a) Video diffusion models exhibit hierarchical denoising: high-noise stages ($t=0.9, 0.6$) establish motion and composition, while low-noise stages ($t=0.3, 0.0$) refine details and textures. (b) Gradient analysis under quality degradations, averaged over 120 samples with three VQ degradation types (blur, compression, noise) and MQ degradation (shuffle). VQ degradation aligns with the original at high timesteps, while MQ degradation aligns at low timesteps, revealing that quality-imbalanced samples can match golden data's gradients at appropriate timesteps and should be strategically allocated across the denoising process.}
    \label{fig:analysis}
\end{figure*}

\subsection{The Vision-Motion Quality Dilemma}
\label{subsec:dilemma}
We employ the VideoAlign~\citep{videoalign} framework to score a random sampled subset of the Koala-36M dataset~\citep{koala}. Since the original VideoAlign is trained on generated videos, we retrain the model on real-world videos to better suit our analysis of natural video distributions. We then plot the distribution of MQ and VQ scores.
As illustrated in Fig.~\ref{fig:intro}(b), we divide the data into four quadrants based on the median values of the collected data for analysis purposes (note that practical filtering pipelines would use much higher thresholds). Notably, data satisfying both high VQ and high MQ (above median) accounts for only 21.9$\%$, while high-VQ/low-MQ and high-MQ/low-VQ data each comprise around 28$\%$. This reveals a clear pattern: data excelling in both metrics is less common than data showing strength in one metric while being weak in the other.

In addition, we compute the Pearson correlation coefficient between VQ and MQ as $-0.2187$, with a $p$-value $< 0.001$, indicating a trade-off between VQ and MQ. We term this phenomenon the Motion-Vision Quality Dilemma, which contributes to the scarcity of golden data with high quality in both metrics. Through this analysis, we highlight that while golden data is sparse, there exist large amounts of data that excel in only one metric (either VQ or MQ, but not both). Conventional filtering pipelines~\citep{wan, hunyuanvideo} discard these, yet our timestep selection framework demonstrates their utility in training effective video diffusion models.

\subsection{Quality Dynamics Analysis in Video Diffusion Process}
\label{subsec:analysis}

Prior work has shown that image diffusion models determine layout in early timesteps and refine details in later ones~\citep{painter}. Building upon this insight, we investigate its extension to video diffusion models and empirically verify that they prioritize motion information in early timesteps and visual details in later ones. As depicted in Fig.~\ref{fig:analysis}(a), visualizing the video generation process demonstrates that layout and motion are established early, with content refined subsequently.


Furthermore, to demonstrate that imbalanced data can provide gains comparable to golden data at specific timesteps, we conduct a gradient-based analysis across 120 samples with three types of VQ degradation (blur, compression, noise; 40 each) and MQ degradation (frame shuffling). For each degraded sample, we compute gradients of a video diffusion model~\cite{wan} at various timesteps and measure the L2 distance to gradients from the original sample, averaging across all samples. As shown in Fig.~\ref{fig:analysis}(b), VQ degradation maintains gradient similarity with the original at high timesteps (high noise levels), suggesting that high-MQ/low-VQ data effectively supports motion learning in early denoising stages. Conversely, MQ degradation preserves gradient similarity at low timesteps (low noise levels), indicating that high-VQ/low-MQ data contributes meaningfully to detail refinement in later stages. These complementary patterns motivate our timestep-aware training strategy: samples with high VQ can be effectively utilized at lower timesteps, while samples with high MQ contribute meaningfully at higher timesteps, enabling efficient exploitation of data with varying quality characteristics.

\subsection{Timestep-Selective Training with TQD}
\label{subsec:tqd}

\begin{figure*}
    \centering
    \includegraphics[width=0.93\linewidth]{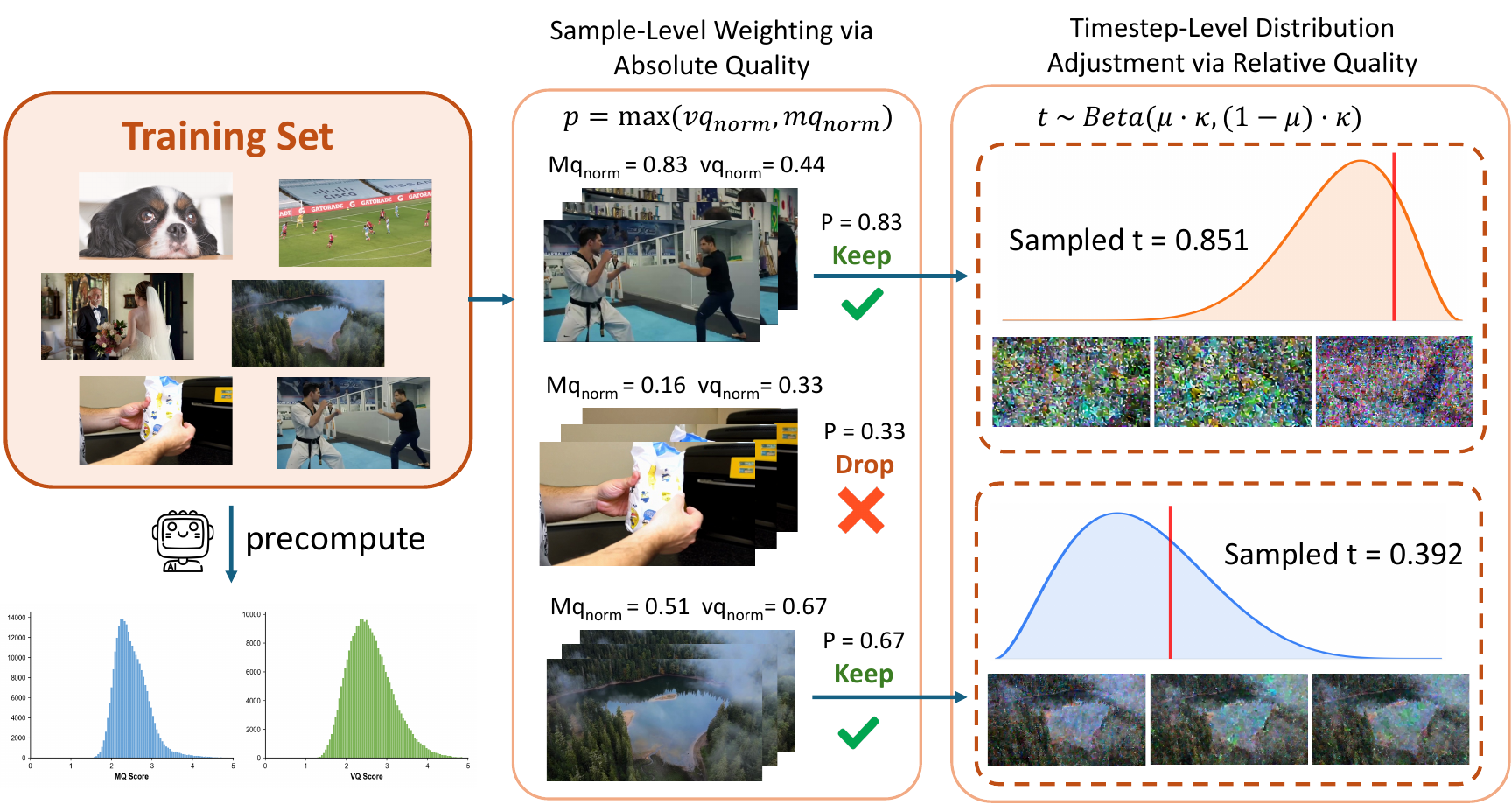}
    \caption{Overview of our timestep selective training process. Given bimodal quality distributions in training data (left), TQD employs quality-based sample dropout (middle) and adaptive timestep sampling via Beta distributions (right). Samples with high motion intensity are directed toward large timesteps, those with high visual quality toward small timesteps, achieving specialized learning across denoising process.}
    \label{fig:method}
\end{figure*}

Based on the analysis in Sec.~\ref{subsec:analysis}, we propose Timestep-aware Quality Decoupling (TQD), a data-aware timestep selection training method that adaptively adjusts sampling strategies for each individual sample. Unlike conventional video generation models that apply uniform or logit-normal timestep sampling equally across all samples, TQD adapts the timestep sampling to each sample's quality characteristics through a two-level adaptive mechanism: (1) sample-level weighting based on absolute quality to down-weight samples lacking distinctive strengths in either dimension, and (2) timestep-level distribution adjustment based on relative quality to match specialized data to appropriate denoising stages. The overall method is shown in Fig.~\ref{fig:method}.

We precompute VQ and MQ scores for each sample in the training dataset, and normalize them to $\text{vq}_{\text{norm}}, \text{mq}_{\text{norm}} \in [0,1]$ via min-max normalization. These normalized scores serve as the foundation for both sample-level and timestep-level adaptation.

\noindent\textbf{Sample-Level Weighting via Absolute Quality.}
To filter low-quality samples where both motion and visual quality are poor, we introduce a quality-based dropout mechanism. We use $\max(\text{vq}_{\text{norm}}, \text{mq}_{\text{norm}})$ as an absolute quality indicator, representing the sample's best attribute. During batch preparation, each sample is retained with probability:
\begin{equation}
p_{\text{sample}} = \max(\text{vq}_{\text{norm}}, \text{mq}_{\text{norm}}).
\label{eq:scaling}
\end{equation}
This weighting scheme ensures that samples with at least one distinctive quality (high MQ or high VQ) receive sufficient training emphasis, while suppressing samples lacking notable strengths in either dimension. Samples with both low MQ and low VQ are naturally down-weighted, preventing them from dominating the training despite having small relative quality differences.

\noindent\textbf{Timestep-Level Distribution Adjustment via Relative Quality.}
For retained samples, we adaptively adjust their timestep sampling distribution based on the relative quality difference between motion and visual attributes. We parameterize the timestep distribution as a quality-dependent Beta distribution over the normalized interval $[0, 1]$:
\begin{equation}
p(t | x_0) \propto \text{Beta}(t; \mu(x_0) \times \kappa(x_0), (1 - \mu(x_0)) \times \kappa(x_0)),
\label{eq:beta_basic}
\end{equation}
where $t \in [0,1]$ denotes the normalized timestep and $x_0$ represents the data sample. The Beta distribution is characterized by two sample-dependent parameters: the center $\mu(x_0)$ and the concentration $\kappa(x_0)$, which are dynamically computed from the sample's quality metrics.

The sampling center $\mu$ captures the relative strength between motion and visual quality:
\begin{equation}
\mu = 0.5 + 0.5 \times (\text{mq}_{\text{norm}} - \text{vq}_{\text{norm}}),
\label{eq:mu}
\end{equation}
where $(\text{mq}_{\text{norm}} - \text{vq}_{\text{norm}}) \in [-1, 1]$ reflects the comparative advantage. Samples with high MQ relative to VQ ($\mu > 0.5$) shift the distribution toward higher timesteps, focusing on motion learning during early denoising stages. Conversely, samples with high VQ relative to MQ ($\mu < 0.5$) bias toward lower timesteps, emphasizing appearance details in later refinement stages.

The concentration parameter $\kappa$ controls the distribution's peakedness based on the quality disparity:
\begin{equation}
\kappa = \kappa_{\text{base}} + (\kappa_{\text{max}} - \kappa_{\text{base}}) \times |\text{mq}_{\text{norm}} - \text{vq}_{\text{norm}}|,
\label{eq:kappa}
\end{equation}
where $|\text{mq}_{\text{norm}} - \text{vq}_{\text{norm}}| \in [0,1]$ quantifies the quality imbalance. A larger disparity produces a higher $\kappa$, yielding a more peaked distribution around $\mu$, thus providing stronger timestep specialization for imbalanced samples. For balanced samples (golden data with high MQ and VQ, or low-quality data already filtered by sample-level weighting), the distribution becomes flatter and approaches the baseline sampling. The base concentration $\kappa_{\text{base}}$ is set to match the original training scheme: $\kappa_{\text{base}} = 2$ for uniform timestep sampling (equivalent to $\text{Beta}(1,1)$), or $\kappa_{\text{base}} = 4$ to approximate logit-normal distributions. This design ensures that when $\text{mq}_{\text{norm}} = \text{vq}_{\text{norm}}$, our method degenerates to the standard sampling strategy.

\noindent\textbf{Combined Quality-Weighted Distribution.}
Combining both sample-level and timestep-level mechanisms, the complete quality-weighted timestep distribution becomes:
\begin{equation}
p(t) \propto \max(\text{vq}_{\text{norm}}, \text{mq}_{\text{norm}}) \cdot \text{Beta}(t; \mu \times \kappa, (1 - \mu) \times \kappa).
\label{eq:beta}
\end{equation}
This formulation ensures that: (1) low-quality samples (LMLV) are down-weighted through the absolute quality term, (2) motion-specialized samples (HMLV) concentrate on large timesteps, (3) visual-specialized samples (LMHV) concentrate on small timesteps, and (4) golden samples (HMHV) maintain balanced timestep coverage. We provide visualization examples of sample-specific timestep distributions in the Appendix for intuitive understanding.

\noindent\textbf{Training Algorithm.}
Our complete training procedure is outlined in Alg.~\ref{alg:tqd}. The method operates in two stages: (1) batch preparation where samples are retained with probability $\max(\text{vq}_{\text{norm}}, \text{mq}_{\text{norm}})$ (Lines 4-10), implementing absolute quality weighting, and (2) timestep sampling where retained samples draw timesteps from Beta distributions parameterized by their relative quality difference (Lines 12-16). This design effectively utilizes imbalanced data while filtering low-quality samples, achieving superior performance to models trained on curated data while expanding the usable training corpus.

\begin{algorithm}[t]
\caption{Timestep-Selective Training with TQD}
\label{alg:tqd}
\begin{algorithmic}[1]
\Require Dataset $\mathcal{D}$ with precomputed $\{\text{vq}, \text{mq}\}$, concentrations $\kappa_{\text{base}}, \kappa_{\text{max}}$
\State Normalize $\{\text{mq}, \text{vq}\}$ to $[0,1]$ for all $x_0 \in \mathcal{D}$  
\State \textit{// Sample-level weighting via absolute quality}
\Function{PrepareBatch}{$\mathcal{D}$, batch\_size}
    \State $\mathcal{B} \leftarrow \emptyset$
    \While{$|\mathcal{B}| <$ batch\_size}
        \State Sample data $x_0$ from $\mathcal{D}$ with probability $\max(\text{vq}_{\text{norm}}, \text{mq}_{\text{norm}})$ \Comment{Eq.~\eqref{eq:scaling}}
        \State Add $x_0$ to $\mathcal{B}$
    \EndWhile
    \State \Return $\mathcal{B}$
\EndFunction
\State \textit{// Timestep-level sampling via relative quality}
\While{not converged}
    \State $\mathcal{B} \leftarrow$ \Call{PrepareBatch}{$\mathcal{D}$, batch\_size}
    \For{each $x_0 \in \mathcal{B}$}
        \State Compute $\mu$ and $\kappa$  \Comment{ Eq.~\eqref{eq:mu}~\eqref{eq:kappa}}
        \State Sample $t \sim \text{Beta}(\mu \kappa, (1-\mu)\kappa)$ \Comment{Eq.~\eqref{eq:beta}}
        \State Sample $x_1 \sim \mathcal{N}(0, I)$
        \State Compute $x_t \leftarrow t x_1 + (1-t) x_0$
    \EndFor
    \State Update model: $\mathcal{L} = \mathbb{E}_{x_0, x_1, t}[\| v_\theta(x_t, t, c) - (x_1 - x_0) \|^2]$
\EndWhile
\end{algorithmic}
\end{algorithm}

\section{Experiments}

\subsection{Implementation Details}

\subsubsection{Model Setup}
We conduct experiments on two high performance video diffusion models: CogVideoX-5B~\cite{yang2024cogvideox} and Wan-T2V-1.3B~\cite{wan}. CogVideoX employs a standard diffusion process with uniform timestep sampling, while Wan-T2V utilizes flow matching with logit-normal timestep sampling. This diverse setup demonstrates the robustness of our method across different generative frameworks and sampling strategies.

\subsubsection{Dataset Setup}
We curate a training set of 280K video-text pairs online for supervised fine-tuning. All samples are scored using VideoAlign~\cite{videoalign} to obtain MQ and VQ metrics. The quality distribution is relatively balanced across the dataset. To analyze the effectiveness of our method under different quality regimes, we partition the data into four categories based on quality thresholds (MQ=2.5, VQ=2.7): High-MQ-High-VQ (HMHV, 100K samples), High-MQ-Low-VQ (HMLV, 80K samples), Low-MQ-High-VQ (LMHV, 80K samples), and Low-MQ-Low-VQ (LMLV, 30K samples). Detailed quality distributions are provided in the Appendix. Dense captions for all videos are generated using a vision-language model~\cite{gemini}.

\subsubsection{Training Setup}
We finetune Wan-T2V-1.3B with full parameter training using a batch size of 128, and fine-tune CogVideoX-5B with LoRA (rank=256)~\cite{lora} using a batch size of 96. All models are trained for 2,000 steps from their respective base checkpoints. To comprehensively evaluate our method, we construct three training configurations: Set-A uses all four quality categories to simulate realistic data distributions commonly encountered in practice; Set-B uses only HMLV and LMHV samples to assess our method's effectiveness under extreme quality imbalance where dual-quality data is completely absent; Set-C uses only HMHV samples to verify whether TQD provides additional gains even on rare golden data.

\subsection{Quantitative Evaluation}

We evaluate video generation quality using VBench~\cite{vbench} and VideoAlign~\cite{videoalign}. Following standard practice, we use an LLM~\cite{gemini} to expand evaluation prompts into detailed descriptions matching our training data format. The baseline methods employ the default fine-tuning strategies for each model: uniform timestep sampling for CogVideoX and logit-normal sampling for Wan.

As shown in Table~\ref{tab:main_results}, TQD yields consistent gains across all data scenarios. Notably, under extreme quality imbalance (Set-B containing only HMLV and LMHV samples), TQD reaches or even surpasses results from naive fine-tuning on naturally distributed data—for example, Wan-T2V with TQD on Set-B (MQ: 2.1477, VQ: 3.2679) exceeds the baseline on Set-A (MQ: 2.1388, VQ: 3.2537). This shows that timestep-aware training can effectively exploit separated imperfect data and close, or even reverse, the gap with mixed-quality training. TQD also improves performance when trained solely on high-quality data (Set-C), demonstrating its robustness across different data regimes. We also observe smaller gains on CogVideoX, likely due to its LoRA-based fine-tuning setup, which constrains the model’s ability to fully adapt.

\renewcommand{\arraystretch}{0.98}
\begin{table*}[t]
\centering
\caption{Quantitative comparison on VBench and VideoAlign metrics across different data settings. Best results in each setting are shown in \textbf{bold}.}
\label{tab:main_results}
\resizebox{\textwidth}{!}{
\begin{tabular}{@{}ccc|cccc|cc@{}}
\toprule

\textbf{Model} & \textbf{Setting} & \textbf{Method} & \multicolumn{4}{c|}{\textbf{VBench}} & \multicolumn{2}{c}{\textbf{VideoAlign}} \\
\cmidrule(lr){4-7} \cmidrule(lr){8-9}
& & & IQ & Aesthetic & Dynamic & MotionSmooth & MQ & VQ \\
\midrule
\multirow{7}{*}{\textbf{Wan 1.3B}} 
& - & Basemodel & 0.6935 & 0.5874 & 0.5938 & 0.9889 & 2.1993 & 3.3236 \\
\cmidrule{2-9}
& \multirow{2}{*}{Set-A} & Baseline & 0.6916 & 0.5722 & 0.5312 & \textbf{0.9935} & 2.1388 & 3.2537 \\
& & TQD & \textbf{0.7010} & \textbf{0.5757} & \textbf{0.6384} & 0.9923 & \textbf{2.2557} & \textbf{3.3450} \\
\cmidrule{2-9}
& \multirow{2}{*}{Set-B} & Baseline & 0.6769 & \textbf{0.5702} & 0.5224 & 0.9923 & 2.0905 & 3.2338 \\
& & TQD & \textbf{0.6921} & 0.5668 & \textbf{0.5447} & \textbf{0.9935} & \textbf{2.1477} & \textbf{3.2679} \\
\cmidrule{2-9}
& \multirow{2}{*}{Set-C} & Baseline & 0.6939 & 0.5798 & 0.5268 & 0.9920 & 2.1917 & 3.3378 \\
& & TQD & \textbf{0.7556} & \textbf{0.5847} & \textbf{0.6473} & \textbf{0.9931} & \textbf{2.2200} & \textbf{3.3743} \\
\midrule
\multirow{7}{*}{\textbf{CogVideoX 5B}} 
& - & Basemodel & 0.6796 & 0.5361 & 0.7455 & 0.9780 & 2.2134 & 3.2379 \\
\cmidrule{2-9}
& \multirow{2}{*}{Set-A} & Baseline & 0.6811 &  \textbf{0.5388} & 0.7500 & \textbf{0.9781} & 2.1855 & \textbf{3.2593} \\
& & TQD & \textbf{0.6816} & 0.5386  & \textbf{0.7723} & 0.9780 & \textbf{2.2345} & 3.2570 \\
\cmidrule{2-9}
& \multirow{2}{*}{Set-B} & Baseline & 0.6798  & 0.5379 & 0.7679 & 0.9781 & 2.1821 & 3.2407 \\
& & TQD & \textbf{0.6848}  & \textbf{0.5381} & \textbf{0.7768} & \textbf{0.9793} & \textbf{2.2010} & \textbf{3.2499} \\
\cmidrule{2-9}
& \multirow{2}{*}{Set-C} & Baseline & 0.6814  & 0.5386  & \textbf{0.7723} & 0.9779 & 2.2028 & 3.2373 \\
& & TQD & \textbf{0.6857}  & \textbf{0.5397}  & 0.7679 & \textbf{0.9783} & \textbf{2.2448} & \textbf{3.2392} \\
\bottomrule
\end{tabular}
}
\end{table*}
Interestingly, we also observe that fine-tuning with our curated dataset enhances downstream physical reasoning capabilities. We evaluate Wan 1.3B Set-A checkpoints on VideoPhy2~\cite{videophy2}, specifically testing on hard cases that require understanding of physical dynamics. As shown in Table~\ref{tab:physical}, TQD surpasses the baseline in both physical commonsense adherence and semantic alignment. To isolate TQD's contribution from data filtering effects, we include a Data Filter Only variant that applies quality-based dropout without adaptive timestep allocation. TQD outperforms both, confirming that gains stem from timestep-aware routing rather than mere data selection.
\begin{table}[h]
\centering
\caption{Quantitative comparison on Subject Alignment (SA) and Prompt Consistency (PC) metrics with VideoPhy2.}
\label{tab:physical}
\setlength{\tabcolsep}{12pt}
\begin{tabular}{@{}lcc@{}}
\toprule
\textbf{Model} & \textbf{SA} $\uparrow$ & \textbf{PC} $\uparrow$ \\
\midrule
Wan 1.3B & 3.738 & 3.690 \\
Baseline & 3.786 & 3.709 \\
Data Filter Only & 3.795 & 3.756 \\
TQD (Ours) & \textbf{3.820} & \textbf{3.843} \\
\bottomrule
\end{tabular}
\end{table}

\subsection{Qualitative Evaluation}

\begin{figure*}
    \centering
    \includegraphics[width=\linewidth]{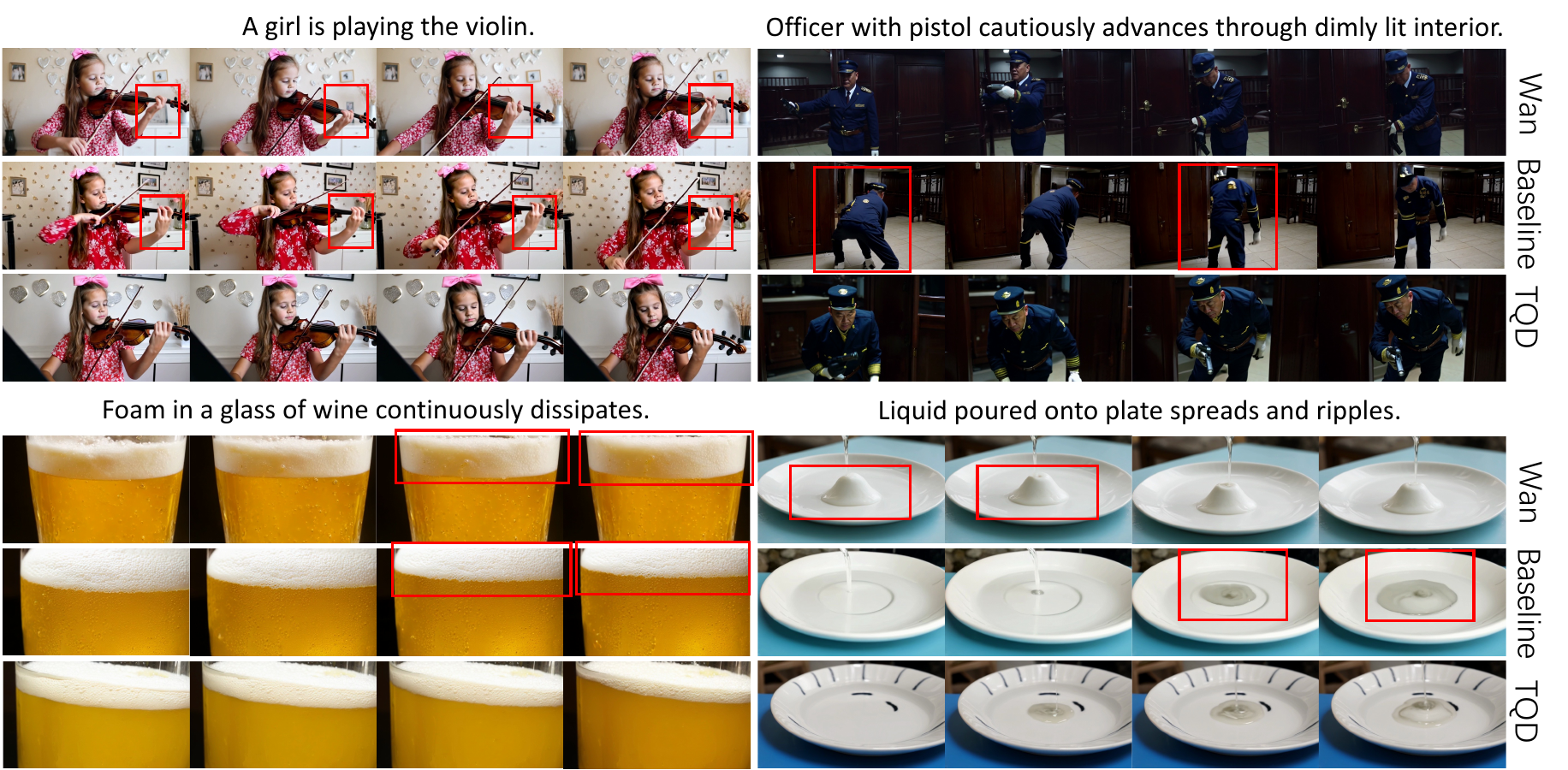}
    \caption{Qualitative comparison. TQD demonstrates superior performance across multiple quality dimensions. Top rows: improved visual quality with reduced hand distortion (left) and better motion coherence matching prompt dynamics (right). Bottom rows: enhanced physical plausibility in foam dissipation (left) and liquid dynamics (right), as highlighted in red boxes.}
    \label{fig:qualitative}
\end{figure*}

Fig.~\ref{fig:qualitative} compares models trained on quality-imbalanced data from Set-B. Baseline training on such imbalanced data revealsquality degradation: visual artifacts such as hand distortions (top-left), motion incoherence failing to match prompt dynamics (top-right), and struggles with physical realism including unnatural foam dissipation and improper liquid spreading (bottom rows, red boxes).
TQD effectively addresses these limitations by directing each sample to timesteps that match its quality strengths—motion-rich samples to high-noise stages, visually clean samples to low-noise stages. This yields videos with both visual fidelity and motion coherence, while notably improving physical plausibility in dynamics like foam dissipation and liquid flow, achieving quality on par with curated-data training.

\subsection{Ablation Study}
\label{subsec:ablation}

We conduct comprehensive ablation studies to validate our design choices. All experiments are performed on Wan 1.3B with Set-A.

\begin{table*}[h]
\centering
\caption{Component ablation of TQD on Wan Set-A. We VBench metrics alongside VideoAlign scores to dissect motion-visual synergies. }
\label{tab:ablation_component}
\renewcommand{\arraystretch}{0.92} 
\resizebox{0.8\linewidth}{!}{
\begin{tabular}{@{}p{3.5cm}|cccc|cc@{}}
\toprule
\centering\textbf{Method} & \multicolumn{4}{c|}{\textbf{VBench}} & \multicolumn{2}{c}{\textbf{VideoAlign}} \\
\cmidrule(lr){2-5} \cmidrule(lr){6-7}
\centering & IQ  & Aesthetic & Dynamic  & MotionSmooth  & MQ & VQ \\
\midrule
Baseline 
& 0.6916 & 0.5722 & 0.5312 & 0.9935 & 2.1388 & 3.2537 \\
\quad + Adaptive timestep 
& 0.6972 & 0.5735 & 0.6340 & \textbf{0.9937} 
& 2.2193 & 3.3044 \\
\quad + Quality dropout 
& 0.6956 & 0.5701 & 0.5357 & 0.9916 
& 2.1909 & 3.2921 \\
\quad + Both (TQD) 
& \textbf{0.7010} & \textbf{0.5757} & \textbf{0.6384} & 0.9923 & \textbf{2.2557} & \textbf{3.3450} \\
\bottomrule
\end{tabular}
}
\end{table*}
\noindent\textbf{Component Analysis.}
Table~\ref{tab:ablation_component} dissects TQD's two core components on Wan Set-A. Adaptive timestep allocation via Beta distributions brings substantial improvements by directing samples to their optimal denoising stages based on quality characteristics. Quality-based dropout further enhances performance by filtering low-quality samples and rebalancing the training distribution. Combining both mechanisms yields synergistic gains (+0.1169 MQ, +0.0913 VQ), demonstrating that sample-level filtering and timestep-level adaptation mutually reinforce each other to achieve superior video quality. MotionSmooth scores shows minimal and comparable variation across methods.

\noindent\textbf{Concentration Parameter.}
The concentration parameter $\kappa_{\text{max}}$ controls the peakedness of timestep distributions—higher values create more concentrated distributions, with samples exhibiting larger MQ-VQ disparities receiving proportionally higher $\kappa$ for stronger specialization. We set $\kappa_{\text{base}} = 2$ for CogVideoX to match its baseline uniform sampling (Beta(1,1)), and $\kappa_{\text{base}} = 4$ for Wan to approximate its logit-normal sampling. Table~\ref{tab:ablation_kappa} systematically examines $\kappa_{\text{max}}$ across both architectures. We observe that CogVideoX achieves optimal performance at lower values, while Wan performs best at higher values. This difference suggests that models trained with uniform sampling may require more conservative distribution adjustments. Importantly, while $\kappa_{\text{max}}$ influences performance, all configurations of Wan substantially outperform their respective baselines (Wan baseline: MQ 2.1388, VQ 3.2537), demonstrating TQD's robustness to this hyperparameter. Based on these findings, we adopt $\kappa_{\text{max}} = 20$ for CogVideoX and $\kappa_{\text{max}} = 30$ for Wan in all our experiments.

\begin{table}[h]
\centering
\caption{Ablation study on concentration parameter $\kappa_{\text{max}}$ across CogVideoX and Wan architectures. }
\label{tab:ablation_kappa}
\setlength{\tabcolsep}{8pt}
\begin{tabular}{@{}c|cc|cc@{}}
\toprule
\multirow{2}{*}{$\kappa_{\text{max}}$} & \multicolumn{2}{c|}{\textbf{CogVideoX}} & \multicolumn{2}{c}{\textbf{Wan}} \\
\cmidrule(lr){2-3} \cmidrule(lr){4-5}
& \textbf{MQ} $\uparrow$ & \textbf{VQ} $\uparrow$ & \textbf{MQ} $\uparrow$ & \textbf{VQ} $\uparrow$ \\
\midrule
10 & 2.2018 & \textbf{3.2596} & 2.2282 & 3.2808 \\
20 & \textbf{2.2345} & 3.2570 & 2.2338 & 3.3020 \\
30 & 2.2249 & 3.2543 & \textbf{2.2557} & \textbf{3.3450} \\
40 & 2.2192 & 3.2535 & 2.2503 & 3.3152 \\
50 & 2.2199 & 3.2504 & 2.2378 & 3.3089 \\
\bottomrule
\end{tabular}
\end{table}

\section{Conclusion}
\label{sec:conclusion}


This paper identifies the Motion-Vision Quality Dilemma—a negative correlation between motion and visual quality that makes golden data statistically rare. Through gradient analysis, we reveal that quality-imbalanced samples can match golden data's learning effectiveness at appropriate timesteps, motivating a shift from data filtering to timestep-aware training. We introduce Timestep-aware Quality Decoupling (TQD), which dynamically aligns each sample's timestep distribution with its quality profile, achieving performance comparable to golden-data-only training with substantial gains on imbalanced datasets.
\clearpage
\setcounter{page}{1}
\maketitlesupplementary

\section{Timestep Distribution Analysis}

\begin{figure}[h]
    \centering
    \includegraphics[width=\linewidth]{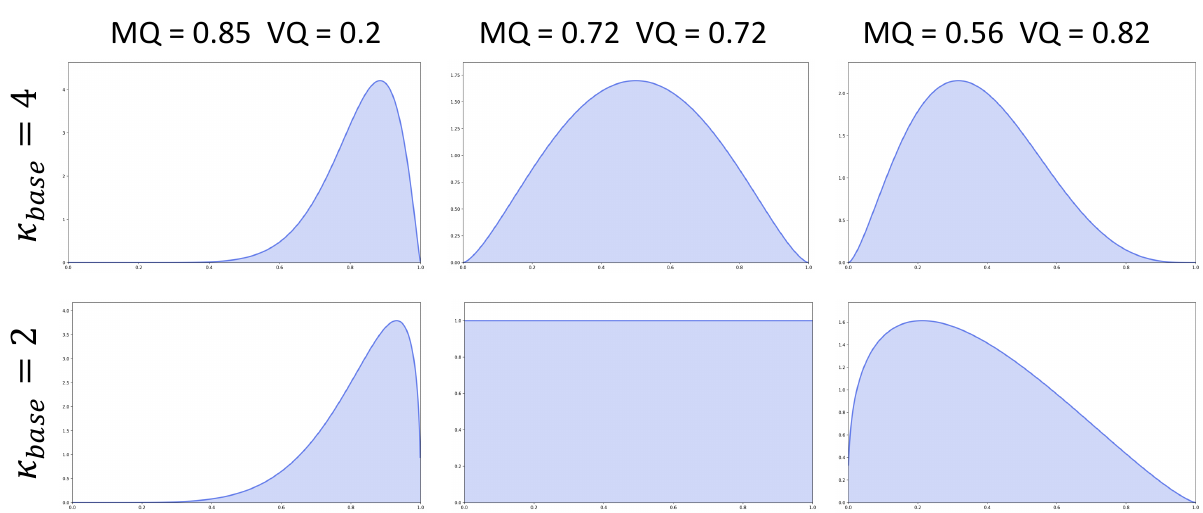}
\caption{
    Timestep sampling distributions under varying quality scores.
    We visualize the Beta distributions with $\kappa_{\text{base}} = 4$ (top) and $\kappa_{\text{base}} = 2$ (bottom) for samples with different MQ and VQ scores.
    Left: High MQ, low VQ shifts toward large timesteps.
    Middle: Equal MQ and VQ yields centered distributions.
    Right: Low MQ, high VQ concentrates on small timesteps.
    Note that $\kappa_{\text{base}} = 2$ degenerates to uniform when MQ = VQ (middle-bottom).}
    \label{fig:timestep}
\end{figure}

Fig.~\ref{fig:timestep} visualizes the timestep distributions across samples with different MQ and VQ scores under two $\kappa_{\text{base}}$ configurations. 
We set $\kappa_{\text{base}} = 2$ and $4$ such that when $\text{MQ} = \text{VQ}$, the distributions degenerate to uniform and logit-normal, respectively.
Notably, samples with higher MQ scores shift the distribution toward larger timesteps, 
while those with higher VQ scores concentrate sampling at lower timesteps. 
This adaptive reweighting allows our approach to dynamically allocate timesteps based on sample-specific quality metrics.
However, samples with uniformly low or high MQ and VQ scores exhibit similar relative quality metrics,
rendering timestep reweighting alone insufficient for differentiation.
To complement this, we incorporate a probabilistic retention strategy during data loading,
where each sample is kept with probability $\max(\text{vq}_{norm}, \text{mq}_{norm})$ (see Sec.~\ref{subsec:tqd}).

\section{Computational Analysis}
Our proposed training strategy requires VQ and MQ scores for each sample, which can be pre-computed offline before training begins. During training, we only perform two lightweight operations: sample dropout and timestep distribution computation based on VQ and MQ scores. The former seamlessly integrates with the dataloader's prefetch mechanism, while the latter incurs negligible computational overhead. Consequently, our method achieves nearly identical GPU training time compared to the baseline, making it practically cost-free in terms of computational resources.

\section{Dataset Settings}
\subsection{Data Distribution Visualization}
\label{Data}

\begin{figure}[h]
    \centering
    \includegraphics[width=\linewidth]{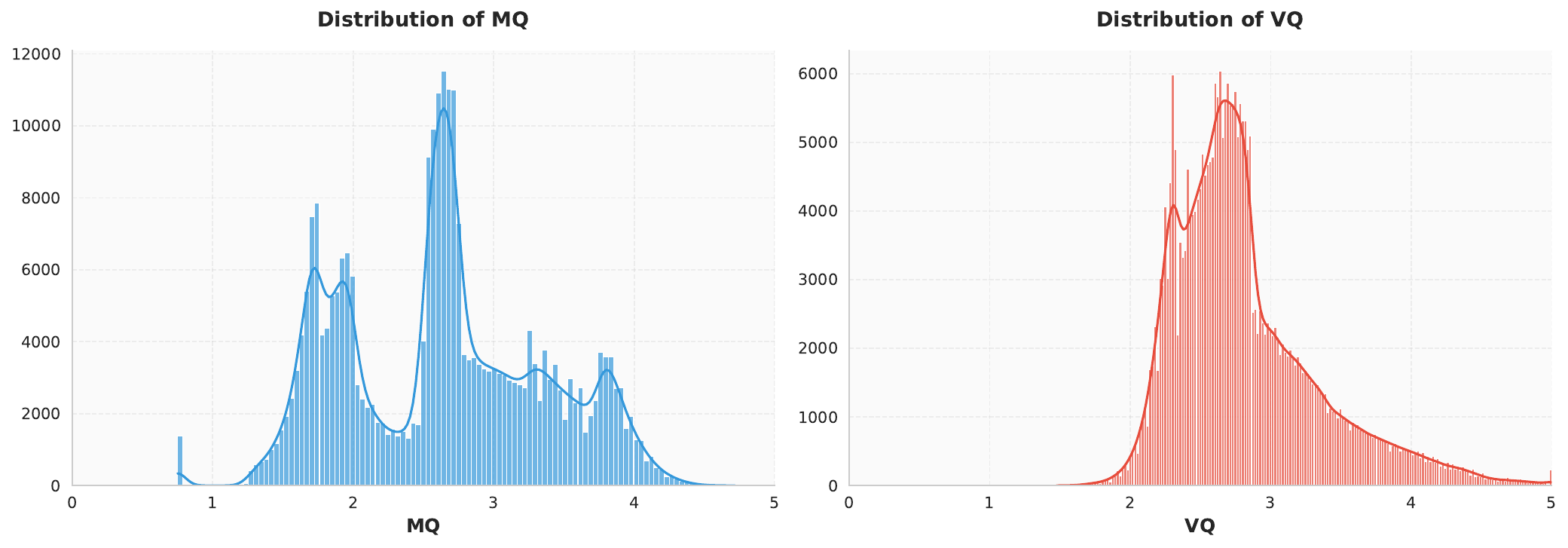}
    \caption{Data distribution of our collected training set.}
    \label{fig:distribution}
\end{figure}

Fig.~\ref{fig:distribution} illustrates the distribution of MQ and VQ scores across our 280K training samples, scored using VideoAlign~\cite{videoalign}. The MQ distribution (left) exhibits clear bimodality with peaks at $\approx$2.3 and $\approx$2.8, reflecting the natural divide between static and dynamic video content. In contrast, VQ distribution (right) follows a unimodal, right-skewed pattern centered at $\approx$2.7, with most samples concentrated between 2.0--3.5. We partition this data into four subsets based on median-aligned thresholds (MQ=2.5, VQ=2.7): HMHV (100K), HMLV (80K), LMHV (80K), and LMLV (20K). Note that these distributions reflect our curated dataset where we deliberately selected different amounts from each quality category and mixed them together to ensure balanced training data.

\subsection{Video Caption Generation}
\begin{figure*}[t]
    \centering
    \includegraphics[width=\linewidth]{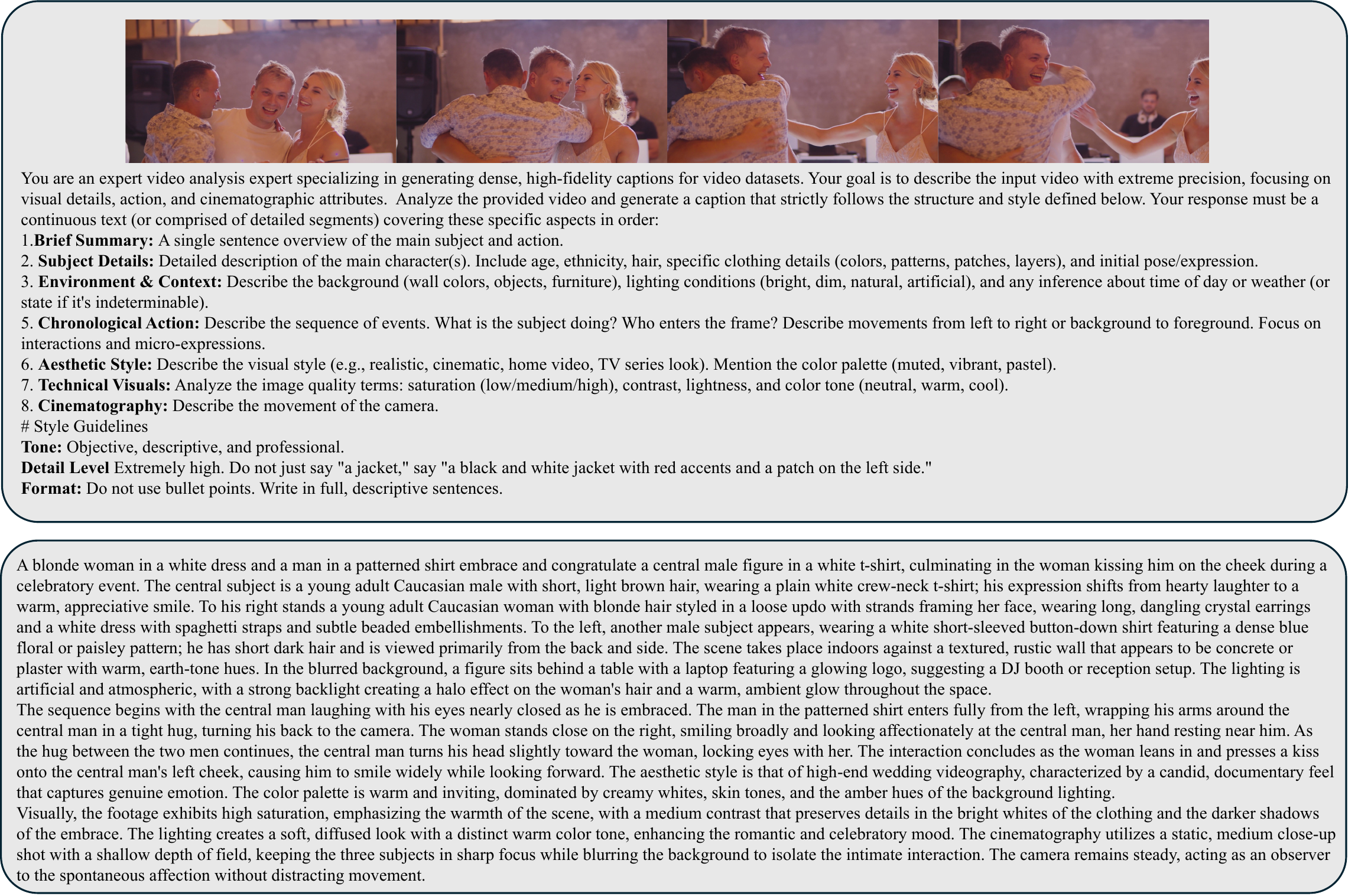}
    \caption{Dense video captioning example. We leverage Gemini 2.5 Pro with structured prompts (top) to generate comprehensive video descriptions. The bottom panel shows an example caption (300+ words) for a wedding scene, demonstrating fine-grained descriptions of subjects' clothing, spatial dynamics, lighting, and camera techniques. This structured approach ensures annotation quality and consistency across our 280K training dataset.}
\label{fig:caption}
\end{figure*}

To obtain precise and detailed video descriptions, we employ Gemini 2.5 Pro~ with carefully designed prompts that emphasize comprehensive scene analysis. Our prompting strategy instructs the model to provide structured captions covering eight key dimensions: (1) brief summary, (2) subject details including appearance and attire, (3) environment and context, (4) chronological actions and interactions, (5) aesthetic style, (6) technical visual attributes, (7) cinematography, and (8) camera movement. This structured approach ensures consistent, high-fidelity annotations that capture both visual content and stylistic nuances essential for training high-quality video generation models. Figure~\ref{fig:caption} illustrates our complete prompting template alongside a representative caption example generated for a wedding celebration scene.

\section{More Qualitative Comparisons}
Additional qualitative video comparisons are included in the supplementary materials,  which further demonstrate  the effectiveness of our approach.

\section{Robustness to Scorer Noise}
\label{sec:scorer_noise}
TQD employs probabilistic Beta-distribution sampling rather than hard-threshold filtering, providing inherent resilience to scorer noise---minor scoring errors only cause slight shifts in the sampling center ($\mu$) without excluding any data. To quantify this, we inject 10\% and 20\% Gaussian noise into MQ and VQ scores on Wan Set-A.

\begin{table}[h]
\centering
\caption{Robustness to scorer noise on Wan Set-A. TQD maintains substantial gains over the baseline even with 20\% noise.}
\label{tab:noise}
\setlength{\tabcolsep}{3pt}
\resizebox{\linewidth}{!}{
\begin{tabular}{@{}l|cccc|cc@{}}
\toprule
\textbf{Method} & \textbf{IQ} & \textbf{Aesthetic} & \textbf{Dynamic} & \textbf{Smooth} & \textbf{MQ} & \textbf{VQ} \\
\midrule
Uniform (no TQD) & 0.6916 & 0.5722 & 0.5312 & 0.9935 & 2.1388 & 3.2537 \\
TQD (0\% noise) & \textbf{0.7010} & \textbf{0.5757} & \textbf{0.6384} & 0.9923 & \textbf{2.2557} & \textbf{3.3450} \\
TQD (10\% noise) & 0.6985 & 0.5743 & 0.6251 & 0.9931 & 2.2431 & 3.3285 \\
TQD (20\% noise) & 0.6953 & 0.5731 & 0.6012 & 0.9928 & 2.2114 & 3.3085 \\
\bottomrule
\end{tabular}
}
\end{table}

As shown in Table~\ref{tab:noise}, TQD with 20\% noise still significantly outperforms the baseline across all metrics. This robustness is further validated by consistent improvements on the independent VBench metrics.

{
    \small
    \bibliographystyle{ieeenat_fullname}
    \bibliography{main}
}


\end{document}